\documentclass[review]{elsarticle}

\usepackage{hyperref}
\usepackage{latexsym} 
\usepackage{graphics}
\usepackage{epsfig}
\usepackage{graphicx}% http://ctan.org/pkg/graphicx
\usepackage{tikz-qtree}
\usepackage{tikz}
\usepackage{amsmath}
\usepackage{dirtytalk} %used for quotations
\usepackage{bbm} %used for the indicator function
\usepackage{algorithm}
\usepackage{pgfplots}
\usepackage{caption}
\usepgfplotslibrary{groupplots}
\usepackage[utf8]{inputenc}
\usepackage[english]{babel}
\usepackage{pgfplots}
\pgfplotsset{compat=1.17}
\usepackage{rotating}
\pgfplotsset{width=11cm,compat=newest}
\usepackage{algorithm}
\usepackage[noend]{algpseudocode}
\makeatletter
\def\BState{\State\hskip-\ALG@thistlm}
\makeatother
\algdef{SE}[DOWHILE]{Do}{doWhile}{\algorithmicdo}[1]{\algorithmicwhile\ #1}%
\usepackage{graphicx}
\usepackage{subfig}
\usepackage{tikz}
\usepackage{pgfplots}

\usepackage{varwidth}
\usepackage{booktabs} % For formal tables
\usepackage{epsfig}
\usepackage{amsfonts}
\usepackage{array}

\usepackage{rotating}
\usepackage{tabularx}
\usepackage{color}
\usepackage{multirow}
\usepackage{url}
\usepackage{lscape}
\usepackage{xargs} 
\newcolumntype{x}[1]{>{\centering\let\newline\\\arraybackslash\hspace{0pt}}p{#1}}

\usepackage{tabularx}
\usepackage{array}
\usepackage{colortbl}
\usepackage{multirow}
\usepackage{datetime}
\usepackage[mathscr]{eucal}
\usepackage{amssymb}
\usepackage{amsmath}
\usepackage{graphicx}
\usepackage{derivative}
\usepackage[utf8]{inputenc}
\usepackage[T1]{fontenc}
\usepackage{mathtools}
\usepackage[thinc]{esdiff}
\usepackage{wrapfig}
\usepackage{caption}
\usepackage{lscape}
\usepackage{rotating}
\usepackage{url}
\usepackage{babel}
\usepackage{translator}
\usepackage{pgfkeys,pgfcalendar}
\usepackage{graphicx}
\usepackage{multirow}
\usepackage{hhline}
\usepackage{diagbox}
\usepackage{eurosym}
\usepackage{amssymb}% http://ctan.org/pkg/amssymb
\usepackage{pifont}% http://ctan.org/pkg/pifont
\graphicspath{ {/home/ciaran/Downloads/} }
\usepackage{acronym}
\usepackage{booktabs}
\def\preparecolorrefs#1{%
  \setcounter{refindex}{0}%
  \whiledo{\value{refindex}<#1}{%
    \stepcounter{refindex}%
    \expandafter\def\csname\therefindex color\endcsname{black}%
  }%
}
\usepackage{hyperref}

\journal{}

\biboptions{authoryear}

\begin{document}

\begin{frontmatter}

\title{Optimizing Quantile-based Trading Strategies in Electricity Arbitrage}
% \tnotetext[mytitlenote]{}

\author{Ciaran O'Connor}
\address{SFI CRT in Artificial Intelligence, School of Computer Science \& IT, University College Cork, Ireland}
\ead{119226305@umail.ucc.ie}

\author{Joseph Collins}
\address{School of Mathematical Sciences, University College Cork, Ireland}
\ead{98718584@umail.ucc.ie}

\author{Steven Prestwich, Andrea Visentin}
\address{Insight Centre for Data Analytics, School of Computer Science \& IT, University College Cork, Ireland}
\ead{s.prestwich@cs.ucc.ie, andrea.visentin@ucc.ie}

\begin{abstract}
Efficiently integrating renewable resources into electricity markets is vital for addressing the challenges of matching real-time supply and demand while reducing the significant energy wastage resulting from curtailments. 
To address this challenge effectively, the incorporation of storage devices can enhance the reliability and efficiency of the grid, improving market liquidity and reducing price volatility. 
In short-term electricity markets, participants navigate numerous options, each presenting unique challenges and opportunities, underscoring the critical role of the trading strategy in maximizing profits.
This study delves into the optimization of day-ahead and balancing market trading, leveraging quantile-based forecasts. Employing three trading approaches with practical constraints, our research enhances forecast assessment, increases trading frequency, and employs flexible timestamp orders. 
Our findings underscore the profit potential of simultaneous participation in both day-ahead and balancing markets, especially with larger battery storage systems; despite increased costs and narrower profit margins associated with higher-volume trading, the implementation of high-frequency strategies plays a significant role in maximizing profits and addressing market challenges. Finally, we modelled four commercial battery storage systems and evaluated their economic viability through a scenario analysis, with larger batteries showing a shorter return on investment.
\end{abstract}


\end{frontmatter}

\section{Introduction}\label{sec:introduction}
The 21st century faces the challenge of incorporating sustainable energy sources into our power systems. European electricity markets have experienced a remarkable surge in variable renewable generation, driven by policy incentives and renewable portfolio standards (\cite{statisticsexplained}). Over the past decade, global adoption of wind and solar power has steadily increased. Ireland experience a fast expansion of the renewable generation share in its electricity mix, rising from 21.7\% in 2014 to 39.5\% in 2022 (\cite{Eirgrid2022Renewables}).  
This widespread integration can introduce volatility in net power supply due to rapid and unforeseen changes in their output, potentially resulting in reliability concerns within the power system (\cite{martinez2016impact}).

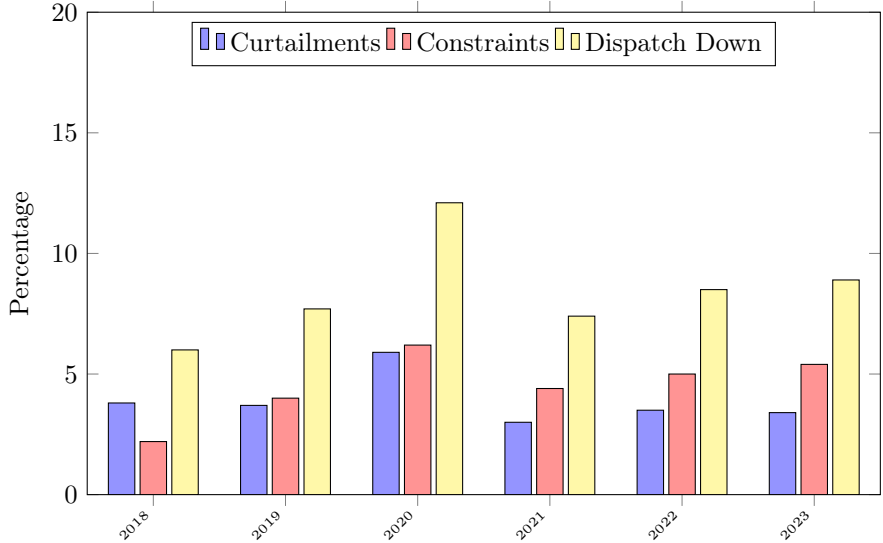
\begin{figure}
    \centering
    \begin{tikzpicture}
        \begin{axis}[
            ybar,
            width=1.0\textwidth,
            height=8cm,
            ymin=0,
            ymax=20,
            ylabel={Percentage},
            symbolic x coords={2018, 2019, 2020, 2021, 2022, 2023},
            xtick=data,
            x tick label style={rotate=45, anchor=east, font=\tiny},
            legend style={at={(0.5,0.98)}, anchor=north, legend columns=-1},
            cycle list={ % Define distinct colors for bars
                {fill=blue!60,draw=black,fill opacity=0.7},
                {fill=red!60,draw=black,fill opacity=0.7},
                {fill=yellow!60,draw=black,fill opacity=0.7},
            }
        ]
        
        % Data for Curtailments, Constraints, Dispatch Down for each year
        \addplot coordinates { (2018, 3.8) (2019, 3.7) (2020, 5.9) (2021, 3.0) (2022, 3.5) (2023, 3.4) };
        \addplot coordinates { (2018, 2.2) (2019, 4.0) (2020, 6.2) (2021, 4.4) (2022, 5.0) (2023, 5.4) };
        \addplot coordinates { (2018, 6.0) (2019, 7.7) (2020, 12.1) (2021, 7.4) (2022, 8.5) (2023, 8.9) };

        \legend{Curtailments, Constraints, Dispatch Down}
        \end{axis}
    \end{tikzpicture}
    \caption{Annual percentages of curtailments, constraints, and dispatch down incidents, illustrating the impact of renewable energy challenges on wind output (\cite{Eirgrid}).}
    \label{curt}
\end{figure}

Curtailments arising from surplus renewable energy generation during low-demand periods persistently impede the expansion of renewable energy capacity. In Ireland, curtailment rates, illustrated in Figure \ref{curt}, range between 3.0\% and 5.9\%, indicating that striving for increased capacity can be counterproductive, imposing significant economic burdens. In a recent study, Ireland had the highest curtailment percentage amongst the countries analysed (\cite{yasuda2022curtailment}). This led to an increase in studies aimed at the facilitation of renewables integration. Many of these were leveraging machine learning; for example, \cite{cardo2024data} developed a machine learning framework to forecast system non-synchronous penetration in the Irish market.

To contribute to the success of this integration, modern electricity markets have turned to Battery Energy Storage Systems (BESS) as a powerful tool to mitigate the volatility in supply and demand caused by fluctuations in renewable energy output. 
BESS offers a compelling solution for reducing or even eliminating these limitations while optimizing the efficient utilization of renewable energy. Strategic BESS deployment can alleviate network constraints, resulting in an overall improvement in grid efficiency.
Moreover, the economic implications of curtailment, particularly the energy wastage caused by inadequate demand, underscore the need for effective solutions. Addressing these challenges enables a more sustainable and efficient energy future.

The rising deployment of BESS underscores the increasing emphasis on energy arbitrage revenues (\cite{timeraenergyBacktestBattery}). BESS enables efficient energy redistribution, exploiting price differentials between low-demand and high-demand periods, thus reducing reliance on expensive fast-responding generation units like thermal peaker plants. Additionally, BESS contributes to load shifting, improved power quality, and grid stability.
In the Irish single electricity market, the Day-Ahead Market (DAM) and the Balancing Market (BM) play pivotal roles in grid stability. The DAM allows trading for electricity delivery the next day through daily auctions, while the BM acts as a real-time mechanism for balancing supply and demand. However, escalating market price volatility, fueled by forecast errors in renewable electricity generation (\cite{do2019impact, maciejowska2020assessing}), especially at the day-ahead stage, presents challenges in maximizing the potential of renewable energy integration (\cite{ortner2019future}).
Effective BESS arbitrage in these markets hinges on accurate probabilistic electricity price forecasting (PEPF), enhancing profitability and risk mitigation (\cite{bunn2018trading}).

\subsection{Motivation and contributions}
In this paper, we respond to recent shifts in research trends that go beyond the traditional focus on PEPF metrics, which prioritize reliability, sharpness, and resolution to include trading strategies to better benchmark the forecasts. Our study delves into trading in the DAM \& BM through probabilistic forecasting, making contributions to addressing gaps in the existing literature; namely:
\begin{itemize}   
    \item We introduce a comprehensive trading approach, incorporating practical BESS constraints, to evaluate the usefulness of PEPF in both markets. We provide a flexible Python implementation that facilitates future battery trading scenario analysis with varied constraints.

    \item We investigate the impact of various fundamental trading constraints on profitability and the assessment of trading strategies, bridging theoretical models with practical market applications.    
    \item  Our study improves trading opportunities with a unique dual-market perspective, employing a real-time forecasting approach that simultaneously focuses on both the DAM and BM with 8-hour BM predictions, aligning DAM with BM positions to optimize trading decisions.  

    \item Our paper extends recent PEPF developments, providing a scenario analysis, assessing battery asset economic viability across four scenarios over a 15-year period, covering various aspects of battery asset economic viability, aiding decision-makers in navigating the evolving cost landscape.

    \item In brief, we explore the potential of quantile forecasts in energy markets, emphasizing battery selection, trading strategies, and economic viability to empower market participants in dynamic electricity markets.
\end{itemize}

The rest of this paper is structured as follows: in Section \ref{pepfliteraturereview}, we refer to recent \& salient DAM and BM PEPF publications. Section \ref{Datasetsec} gives a detailed breakdown of our dataset. Section \ref{meth} provides an overview of our approach, models, specifics of the forecasting problem and of each trading strategy and their constraints. In the experimental results \ref{resultssec}, we present trading strategy performance with a 10-year scenario analysis for four separate batteries. Finally, Section \ref{conclusionsec} concludes the paper with observations and suggested future work.

\section{Literature Review}\label{pepfliteraturereview}
This section surveys the relevant literature regarding electricity price forecasting and trading.
In Section \ref{Proabibilisticliterature}, we conduct an exploration of probabilistic forecasting, dissecting uncertainties prevalent in energy markets and providing a current overview of the field. Moving forward to Section \ref{EnergyStorage}, our focus shifts to the economic evaluation of energy storage systems within electricity markets. Finally, Section \ref{Trading} scrutinizes the intricate interplay of accurate forecasts and pragmatic constraints, shedding light on their pivotal role in the formulation of profitable trading strategies.

\subsection{Probabilistic forecasting}\label{Proabibilisticliterature}
PEPF is a critical component of energy markets, primarily focused on load, price, wind, and solar forecasting. It aids in addressing uncertainties tied to various factors like smart grids, renewable integration, and electric vehicle adoption \cite{khosravi2014closure, khajeh2022applications}. 
In recent studies, best practices and evaluation metrics have been explored extensively. \cite{nowotarski2018recent} investigated various approaches, such as autoregressive models, neural networks, quantile regression averaging, and ensembles. They outline the use of metrics like pinball loss and interval score, emphasizing the importance of reliability, sharpness, and rigorous out-of-sample testing. \cite{marcjasz2020probabilistic} incorporated deep learning models into PEPF, highlighting the superiority of seasonal component models. \cite{tzallas2022probabilistic} introduced a lightweight forecasting model for UK electricity prices. \cite{marcjasz2022distributional} emphasized forecast averaging for enhanced accuracy, particularly with distributional deep neural networks. 
\cite{uniejewski2023smoothing} present a novel PEPF approach, employing smoothed QRA with kernel estimation, outperforming benchmarks in reliability, precision, and economic value. \cite{jiang2024electricity} introduce a nonconvex regularized QRA approach to improve EPF accuracy, particularly in PEPF. \cite{janczura2024expectile} present Expectile Regression Averaging (ERA) as an alternative to traditional quantile-based approaches such as QR and QRA, demonstrating improved accuracy in DAM EPF.

The literature on probabilistic forecasting within the BM remains sparse, largely attributed to challenges in data acquisition. In a study by \cite{klaeboe2015benchmarking}, state-aware models such as SARMA and ARM were compared with purely time series-based models like ARMA and ARX. The results indicated that incorporating contextual information enhanced forecasting accuracy. However, these investigations often feature limited time horizons, typically extending only up to one hour ahead, leaving questions unanswered regarding longer-term forecasting in BMs.
In the context of the Belgian electricity market, \cite{dumas2019probabilistic} introduced a methodology for predicting imbalance prices, leveraging historical data, net regulation volume transition probabilities, and reserve activation information. Additionally, research such as \cite{bunn2020analysis} has delved into the predictability of BM prices using regime-switching models, revealing that BM prices display predictable behavior contrary to efficiency conjectures. These insights suggest potential advantages for market participants in devising forecasting and trading strategies grounded in fundamental econometric relationships.
Moreover, in a study conducted by \cite{lucas2020price} focusing on BM price forecasting in Great Britain, machine learning techniques were employed, highlighting the efficacy of Gradient Boosting and Extreme Gradient Boosting algorithms. On the other hand, \cite{narajewski2022probabilistic} explored BM prediction in the German electricity market, uncovering that simplistic models often outperform sophisticated techniques.

\subsection{Energy Storage}\label{EnergyStorage}
Recent European electricity market reforms prioritize resilient mechanisms amid rising renewable energy integration \cite{zachmann2023design}.
The economic viability of energy storage systems in electricity markets has been examined in various contexts. \cite{staffell2016maximising} assessed the profitability of energy storage systems in the British electricity market, emphasizing the role of reserve services. \cite{tohidi2019stochastic} evaluated revenue from flow battery energy storage in photovoltaic (PV) solar plants, considering uncertainties in electricity prices and PV production. \cite{abramova2021optimal} focused on battery storage trading and optimization for enhanced profitability. \cite{graf2022drives} investigated battery pack degradation, particularly the impact of temperature variations. On average, degradation occurred at 1.55\% per year, with temperature-dependent rates ranging from 1.03\% to 2.00\% annually. \cite{wankmuller2017impact} incorporates degradation's arbitrage impact.

\subsection{Trading}\label{Trading}
In the domain of energy trading, crafting lucrative strategies for energy storage systems critically relies on precise probabilistic forecasts. \cite{krishnamurthy2017energy} performed a comprehensive analysis of trading models tailored for both DAM and BM arbitrage, emphasizing the supremacy of stochastic models amid price uncertainty. Their models, including a Quantity-Only bid model and a Price-Quantity bid model, exhibited diverse levels of profitability and risk.
In a parallel investigation, \cite{l2021optimal} delved into optimal bidding strategies within auction-based electricity markets, specifically for portfolios centered on renewable energy sources. Their research underscored the significance of market impact estimation and the consideration of transaction costs in shaping optimal strategies for risk-neutral traders. The analysis spanned various portfolios, encompassing wind and solar power producers, and showcased substantial revenue enhancements achievable with basic forecasting models. 
More recent developments include \cite{uniejewski2021regularized} with the introduction of Lasso quantile regression averaging (QRA), a method refining point forecasts using quantile regression, and later smoothing QRA \cite{uniejewski2023smoothing}, outperforming benchmarks in quantile-based DAM trading strategies.

In summary, these studies enhance our understanding of effective energy trading methodologies, emphasizing the link between probabilistic forecasting, market dynamics, and optimal bidding strategies. Despite significant progress, gaps remain, particularly in extending probabilistic forecasting to the BM and understanding emerging energy storage technologies. As we move forward, our focus is on addressing these gaps, presenting insights and methodologies to fortify energy trading strategies in both the DAM and BM.

\section{Datasets}\label{Datasetsec}
The data was sourced from \href{https://www.sem-o.com/}{SEMO} \& \href{https://www.semopx.com/market-data/}{SEMOpx}, comprising historical and forward-looking data from 2019 to 2022. The dataset can be categorized into two main types of data: \textit{Historic data} and \textit{Forward/future-looking data}.

% version 1:
% \subsection{Day-ahead Market}\label{day-ahead market}
% In our analysis, we focus on predicting DAM prices, with the forecast horizon extending to the subsequent 24 settlement periods.
% The historical data considered for DAM price prediction includes DAM prices for the previous 168 hours and the wind and demand forecasts for the same interval. We then consider the TSO wind and demand forecasts for the forecasting horizon of 24 settlement periods.

% version 2:
\subsection{Day-ahead Market}\label{day-ahead market}
In our analysis, we focus on predicting DAM prices, with the forecast horizon starting at $t$, extending to the subsequent 24 settlement periods, denoted as:
\begin{align}
    Y_{DAM}=[DAM_{t},...,DAM_{t+23}].
\end{align}
The historical data considered for DAM price prediction includes DAM prices for the previous 168 hours and the wind and demand forecasts for the same interval. We then consider the TSO wind and demand forecasts for the forecasting horizon of 24 settlement periods.

\subsection{Balancing Market}\label{balancingmarket}
For the BM, we predict BM prices for the next 16 open settlement periods. The forecast horizon starts at $t+2$ as at time $t$, the market periods $t$ and $t+1$ are already closed, and adjustments can only be made from $t+2$, denoted as:
\begin{align}
    Y_{BM}=[BMP_{t+2},...,BMP_{t+17}].
\end{align}

\subsubsection*{Historic data}
The historical data used for BM price prediction includes:
\begin{itemize}
    \item \textit{BM Prices}: BM prices from the most recent and available 24 hours.
    \item \textit{BM Volume}: The most recent 48 observations of BM volume.
    \item \textit{Forecast Wind - Actual Wind}: Difference in forecast and actual wind data for the last 48 settlement periods.
    \item \textit{Interconnector Values}: Interconnector flows from the previous 24 hours.
    \item \textit{DAM Prices}: DAM prices from the previous 24 hours, used at an hourly granularity for each half-hour settlement period.
\end{itemize}
\subsubsection*{Forward/future-looking data}
The future-looking data for BM price prediction includes:
\begin{itemize}
    \item \textit{Physical Notifications Volume}: The sum of physical notifications for the forecast horizon.
    \item \textit{Net Interconnector Schedule}: Interconnector schedule for the forecast horizon.
    \item \textit{Renewable Forecast}: TSO renewables forecast for non-dispatchable renewables for the forecast horizon.
    \item \textit{Demand Forecast}: TSO demand forecast for the forecast horizon.
    \item \textit{DAM Prices}: DAM prices for the next 8 hours.
\end{itemize}
The variation in time intervals for historical data is due to availability, limited to the most recent and accessible 48 observations from the data source. For further details on our forecasting approach, market structure, datasets, and variables for both the DAM and BM, please refer to \cite{OCONNOR2024101436}.

\section{Methodology}\label{meth}
In this section, we outline the methodology used in our study to investigate the development of BESS trading strategies. These strategies encompass the optimization, and in some cases, co-optimization of revenue streams derived from participation in ancillary service markets, energy-only markets, or both. Just as ancillary service markets contribute to grid stability and reliability, BESS play an increasingly vital role in integrating renewable energy into the evolving energy landscape. To achieve this objective, our primary focus lies in the formulation of trading strategies aimed at maximizing profits through energy arbitrage in the DAM and BM.

\subsection*{Methodology Overview}
In our analysis the forecasting \& trading approach can be broadly viewed as comprising of 5 key steps for both the DAM and BM. That is:
\begin{enumerate}
\item \textit{Data Collection and Preparation:} We gather historical and forward-looking data from the ISEM relating to the DAM and the BM.

\item \textit{Data Pre-processing and Model Optimization:} In preparation for analysis, we pre-process the data. Each predictive model undergoes hyperparameter tuning for each 3-month data subset to enhance its performance.

\item \textit{Walk-Forward Model Validation:} For the reliability of our models, we employed a walk-forward validation process, iteratively updating the time horizon.

\item \textit{Quantile Forecasting:} We generated quantile forecasts using our optimized models based on unseen test data. These forecasts were designed for time horizons of 24 hours for the DAM and 8 hours for the BM.

\item \textit{Trading Strategy Execution:} With quantile predictions in hand, we implemented our chosen trading strategy. This strategy involved submitting quantity bids and offers for positions in both the DAM and BM, aligning with our 24 and 8-hour forecasts.
\end{enumerate} 
This approach involves the strategic purchase of electricity during periods of low predicted prices, subsequent storage, and timely resale during periods of high predicted prices, with the goal of optimising financial returns.

\subsection{Quantile Regression Models}\label{modelss}
All models in our study employ quantile regression, a statistical technique for estimating conditional quantiles, allowing the estimation of target values at specific quantiles (e.g., 0.1, 0.3, 0.5, 0.7, 0.9). These quantiles are used to define a "quantile pair," which establishes a forecast range, encompassing lower and upper bounds, offering potential values within a specified confidence level. For instance, a quantile pair of 0.1 and 0.9, denoted by $\alpha$ = 0.1 and its complementary quantile $1-\alpha$ = 0.9, defines a forecast range that covers the central 80\% of potential outcomes, providing a nuanced perspective on the distribution of forecast possibilities.

\begin{figure}
\centerline{\includegraphics[width=0.7\linewidth]{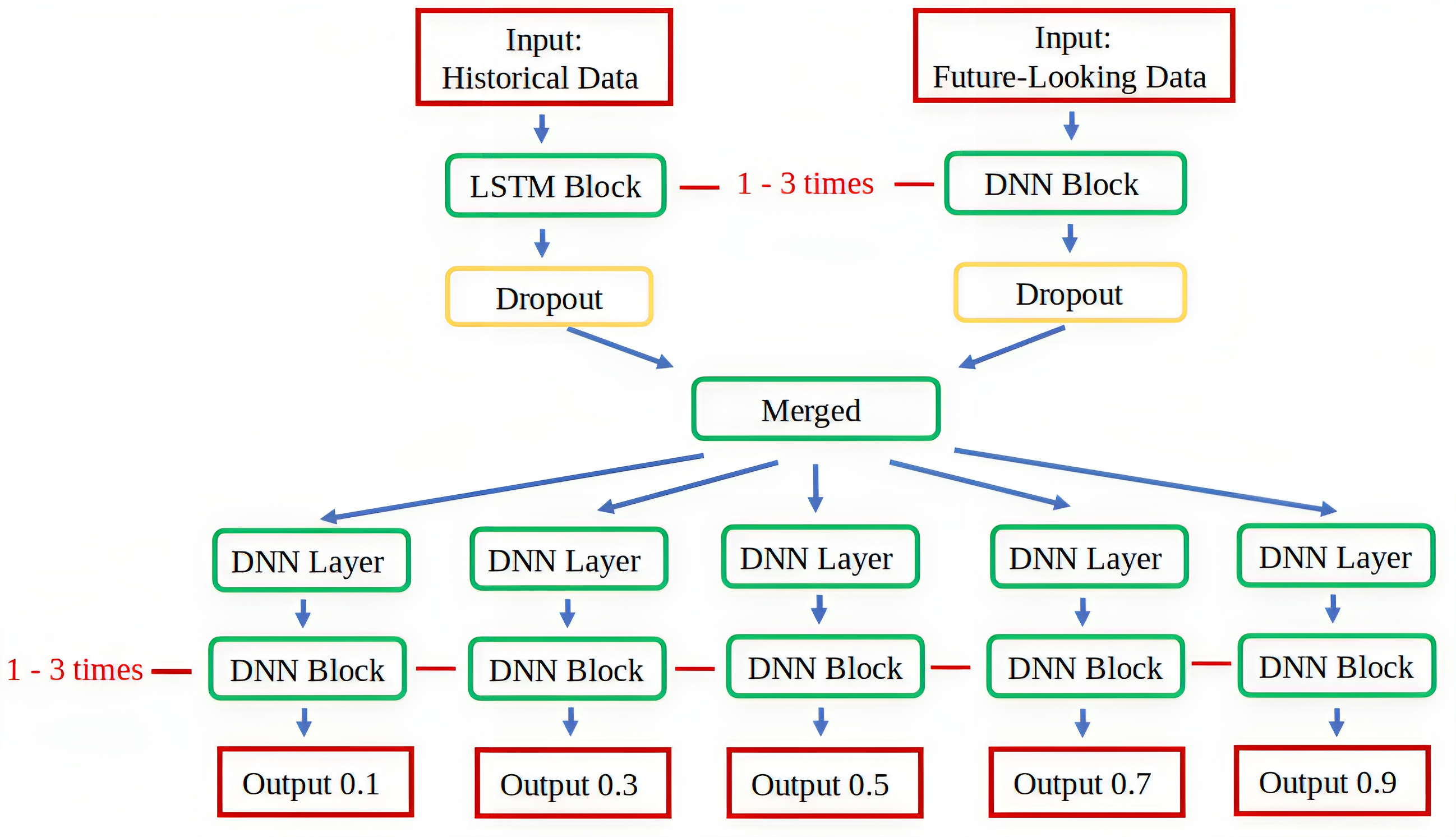}}
\caption{MH RNN DNN Model}
\label{MH_RNNDNN_Q}
\end{figure}

A high-level categorisation of the models includes the following:
\begin{itemize}
    \item \textit{K-Nearest Neighbors} (KNN): A versatile non-parametric algorithm that predicts an instance's output by comparing it to the "K" nearest data points in the feature space, using distance metrics like Euclidean or Manhattan distance. The predicted value is the average of the K nearest data points in the feature space, making it suitable for capturing local patterns in time series data.

    \item \textit{Random Forest} (RF): An ensemble learning method widely utilized in DAM and BM forecasting due to its ability to handle non-linearity and high-dimensional data. By aggregating the predictions of multiple decision trees, RF effectively captures complex temporal relationships and provides robust predictions for both short-term and long-term horizons.

    \item \textit{Light Gradient Boosting Method} (LGBM): Like RF, Gradient Boosting methods have gained popularity in DAM and BM forecasting tasks for their efficiency and scalability. As a boosting algorithm, LGBM sequentially builds decision trees, iteratively improving model performance by focusing on data points with high residuals. This approach enables LGBM to capture intricate temporal patterns and make accurate predictions with minimal computational resources.

    \item \textit{Deep Neural Network} (DNN): A versatile deep learning model equipped with customizable hyperparameters. It adeptly processes both historical data and future projections, facilitating accurate forecasting. Additionally, a hybrid variant merges LSTM and DNN architectures, enhancing predictive capabilities by capturing sequential patterns from historical data while effectively incorporating future trends \cite{lago2018forecasting}. The output layer predicts 8 hours (or 16 timestamps). See Figure \ref{MH_RNNDNN_Q} for visualization.
\end{itemize}

\subsection{Trading Strategies}\label{tradingstrategies}
In this section, we present three trading strategies tailored to optimize profits in electricity markets while considering constraints and price dynamics. These include:
\begin{enumerate}
    \item Trading Strategy 1 (TS1) follows a rule-based heuristic approach, utilizing a straightforward single trade strategy. This strategy ensures that the buy timestamp precedes the sell timestamp.

    \item  Trading Strategy 2 (TS2) maintains TS1's constraints whilst increasing intraday trading frequency with an iterative multi-trade approach.

    \item Trading strategy 3 (TS3) is an optimization-based high-frequency approach that enables flexible buy/sell timestamps while maintaining bottleneck-constrained purchase and sell volumes.    
\end{enumerate}
Our trading strategies are separately applied to both the DAM and the BM. In section \ref{BSCDual}, we present an adapted approach for holding DAM and BM positions concurrently.

\subsection{TS1: Single trade quantile strategy}\label{SingleTrade}
TS1 is a rule-based heuristic trading strategy adopted from \cite{uniejewski2021regularized, uniejewski2023smoothing}, incorporating the same BESS constraints. 
This strategy employs quantile-based forecasts to optimize trading decisions involving a hypothetical battery with capacity $B_{c}$, no discharge limit, discharge efficiency $\mathcal{E}_{d}$, and charge efficiency $\mathcal{E}_{c}$. Over the time horizon of interest (i.e. 1 day for the DAM or 8 hours for the BM), a single buy-sell pair trade is permitted with the requirement that the buy trade occurs before the sell trade.
The key steps for TS1 are as follows:
\begin{enumerate}
    \item \textit{Identify Optimal Timestamps:} Determine timestamps for the minimum and maximum predicted prices for the specified quantile ($\alpha$) and the complementary quantile (1-$\alpha$). Choose the pair with the greater difference.
    \item \textit{Execute Trades:} Submit buy and sell orders at identified timestamps.
    \item \textit{Calculate Profit:} Calculate profit as the difference between the sell and buy prices, adjusted for charge $\mathcal{E}_{c}$ and discharge $\mathcal{E}_{d}$ efficiencies.   
\end{enumerate}
The objective function for the trading strategy is given by:
\begin{equation}
    \max \sum_{t=1}^{T} \left(\mathcal{E}_{d} \cdot \hat{p}_{t_{2}}^\alpha \cdot  - \frac{\hat{p}_{t_{1}}^{1-\alpha} \cdot }{\mathcal{E}_{c}}\right)
\end{equation}
where $\hat{p}_{t}^\alpha$ denotes the model-predicted $\alpha$-quantile, and $\hat{p}_{t}^{1-\alpha} \cdot $ signifies the complementary $(1-\alpha)$-quantile at hour $t$. The total number of time periods, denoted as $T$, corresponds to 24 hourly periods for the DAM and 16 half-hour periods for the BM.

\subsection{TS2: Multi-Trade strategy}\label{MultiTrade}
TS2 extends TS1 by iteratively identifying multiple buy-sell pairs for the 24-hour period in DAM or 8-hour period in BM, maintaining the rule that buy timestamps precede sell timestamps while increasing intraday trading frequency. The first three steps of TS2 are consistent with TS1. The subsequent three steps are specific to TS2:

\begin{enumerate}
    \setcounter{enumi}{3}
    \item \textit{Create Additional Price Subsets:} To adhere to buy/sell constraints, the price data is divided into subsets, enabling the identification of new buy-sell pairs within each subset. The division typically occurs into one subset before the first trade (TS1) and another after it.

    \item \textit{Identify Buy-Sell Pairs:}  Within each subset, as in TS1, identify the timestamps for the minimum and maximum predicted prices for the specified quantile ($\alpha$) and the complementary quantile (1-$\alpha$). For each pair within a subset, submit buy and sell orders for the pair with the greatest difference.
    \item \textit{Calculate Profit:} Calculate profit for each trade pair as the difference between the sell and buy prices, adjusted for efficiency, as in TS1.
\end{enumerate}
While TS2 offers more trading opportunities than TS1, it is limited by the buy/sell order constraint.

\subsection{TS3: High-Frequency Strategy}\label{MVES}
TS3, inspired by \cite{staffell2016maximising}, introduces an optimization-based approach designed to tackle the inherent limitations of battery ramp rates. This challenge is particularly pronounced in larger batteries, where a full charge or discharge within a single trading cycle is typically unfeasible. By incorporating a bottleneck mechanism, TS3 enables greater flexibility in determining buy and sell timestamps, thereby removing the constraint requiring buy timestamps to precede sell timestamps.
The bottleneck within our research encompasses several realistic limitations. The ramp rate $R$ is restricted by factors including the battery's capacity and maximum allowable charging speed, limiting energy acquisition during specific intervals. Similarly, the discharging rate faces constraints linked to the battery's state of charge and the maximum discharging rate, affecting energy release. Battery capacity dictates the volume of energy available for trading. To ensure battery health and reliability, a minimum charge level constraint is in place. Additionally, energy efficiency (denoted as $\mathcal{E}{d}$ and $\mathcal{E}{c}$) causes energy losses during charging and discharging. These limitations define the operational boundaries of our trading strategy and are vital for its performance. 
The key steps of TS3 can be summarized as follows:
\begin{enumerate}
    \item \textit{Find Min and Max Price Periods:} As in TS1  \& TS2, this step involves identifying timestamps for the minimum predicted price associated with a specified quantile ($\alpha$) and the maximum predicted price for the complementary quantile (1-$\alpha$). However, the order of timestamps is now flexible.
    \item \textit{Charging/Discharging Bottleneck:} This step focuses on the charging and discharging bottleneck. When the minimum price period precedes the maximum price period or vice versa, the algorithm dynamically adjusts the charging and discharging operations accordingly to maximize profitability. Further details on this bottleneck and the associated profit calculations are available in Algorithm \ref{bottle}.
    \item \textit{Iterate for Additional Trade Pairs: } The algorithm applies the same process to subsets of price data before, after, and in between each trade pair. This iterative approach allows for the exploration of additional trading opportunities and price subsets.
    \item \textit{Repeat for All Trade Pairs: } Continue iterating through all subsets until no more purchase and sell pairs are available.
\end{enumerate}

\begin{algorithm}[ht!]
\caption{Algorithm 1: TS3 Bottleneck}\label{bottle}
\begin{algorithmic}[1]
\State Initialize variables:  $c$ \textit{(charge level)}, $x_{buy}$ \textit{(charging quantity)}, $x_{sell}$ \textit{(discharging quantity)}, $p_{min}$ \textit{(min price)}, $p_{max}$ \textit{(max price)}, \textit{Profit} 
\State Initialize constants: $R$ \textit{(ramp rate)}, $B_{c}$ \textit{(battery capacity)}, $C_{min}$ \textit{(min charge level)}
\State \textbf{Charging/Discharging Bottlenecks}
\If{$\text{Period of $p_{min}$} < \text{Period of $p_{max}$}$}
    \State \textbf{Charging Strategy:}
    \State $x_{buy} \gets \min(B_{c} - c,\textit{ } R)$ \Comment{Charging Bottleneck}
    \State $c \gets c + x_{buy}$
    \State $x_{sell} \gets \min(c - C_{min},\textit{ }R)$ \Comment{Discharging Bottleneck}
    \State $c \gets c - x_{sell}$
    \State $\text{Profit} \gets (p_{max} \times x_{sell} \times \mathcal{E}_{d}) - (\frac{p_{min} \times x_{buy}}{\mathcal{E}_{c}})$
\Else
    \State \textbf{Discharging Strategy:}
    \State $x_{sell} \gets \min(c - C_{min},\textit{ }R)$ \Comment{Discharging Bottleneck}
    \State $c \gets c - x_{sell}$
    \State $x_{buy} \gets \min(B_{c} - c,\textit{ }R)$ \Comment{Charging Bottleneck}
    \State $c \gets c + x_{buy}$
    \State $\text{Profit} \gets (p_{max} \times x_{sell} \times \mathcal{E}_{d}) - (\frac{p_{min} \times x_{buy}}{\mathcal{E}_{c}})$
\EndIf
\end{algorithmic}
\end{algorithm}
The trading strategy outlined addresses these limitations by dynamically adjusting charging and discharging operations based on the order of minimum and maximum price periods, aiming to maximize profits under the constraints imposed by the bottleneck, thus significantly boosts trading frequency.

\subsubsection*{TS3-Dual: Dual Markets}\label{BSCDual}
Our dual markets approach, \textit{TS3-Dual}, optimizes trading decisions in both the DAM and BM concurrently, ensuring synchronization between them. This strategy capitalizes on the differences in granularity and lead times, such as:
\begin{itemize}
    \item DAM positions are determined at time $t$ for the time range of $t+12$ to $t+36$, with hourly granularity, covering a 24-hour period.
    \item BM positions, in contrast, are generated at time $t+12$, 12 hours after DAM positions have been established. These BM positions are based on an 8-hour ahead forecast extending to $t+20$, with half-hourly granularity.
\end{itemize} 
The emphasis in TS3-Dual is on midday trading opportunities in the DAM, while the remaining time before and after DAM trades is allocated to the BM, capitalizing on BM's consistent volatility.
To achieve this, we create subsets of price data to optimize trades in both the DAM and BM, divided as follows:
\begin{itemize}
    \item Early-Bidding in the BM (PS1): Tailored for BM trades, PS1 focuses on time periods occurring before any DAM trades. It provides an opportunity for strategic planning and early bidding to optimize trading decisions in the BM market.
    
    \item DAM Trades - PS2: Dedicated to the DAM, this subset covers the time range starting from the initial DAM trade pair's minimum period to the maximum period. 
        
    \item Late-Bidding in the BM - PS3: Similar to PS1, PS3 serves BM trades but encompasses periods after DAM trades. It is strategically positioned to accommodate late bidding activities in the BM.
\end{itemize}
We follow the outline for TS3 in each subset, iterating through all subsets until no more purchase and sell pairs are available. Organizing price data into these subsets enables TS3-Dual to efficiently manage trading opportunities in both the DAM and BM while ensuring synchronization between them. Code for all trading strategies can be found at 
\href{https://anonymous.4open.science/r/DAM-BM_Trading-2B30/}{GitHub}\footnote{\url{https://anonymous.4open.science/r/DAM-BM_Trading-2B30/}}

\section{Experimental Results}\label{resultssec}
In this section, we present the results of our study, focusing on the practical applications of quantile forecasts in BESS trading within the DAM and BM. 
Our analysis evaluates three distinct trading strategies: TS1, TS2, and TS3. Each incorporates quantile forecasts with diverse trading constraints to optimize efficiency. 
Subsequent sections offer a detailed exploration of these strategies, encompassing quantile pair selection, model choice, forecast accuracy, and highlighting key distinctions between the DAM and BM that influence trading outcomes.
Furthermore, we conduct a thorough BESS economic analysis, highlighting the economic implications of these strategies within the DAM and BM.

\subsection{Trading strategies} 
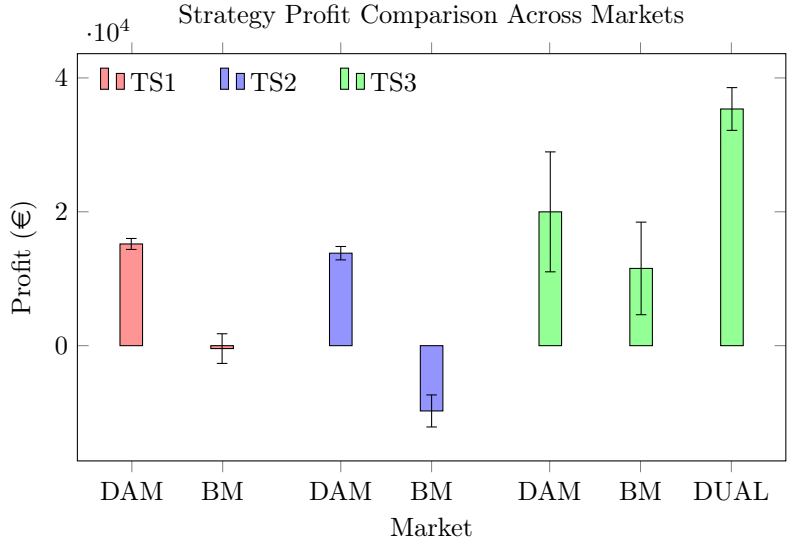
\begin{figure}[ht!]
    \centering
    \begin{tikzpicture}
        \begin{axis}[
            ybar,
            bar width=0.3cm, % Adjust the bar width for the first two bars
            ylabel={Profit (\euro)},
            xtick={0.5, 3.0, 6.25, 8.75, 12.0, 14.5, 17.0},
            xticklabels={DAM, BM, DAM, BM, DAM, BM, DUAL},
            xticklabel style={text width=1.5cm, align=center},
            enlarge x limits=0.15,
            title={Strategy Profit Comparison Across Markets},
            xlabel={Market},
            legend style={at={(0.02,0.98)},anchor=north west,legend columns=-1,legend style={draw=none},/tikz/every even column/.append style={column sep=0.5cm}},
            cycle list={ % Define distinct colors for bars
                {fill=red!60,draw=black,fill opacity=0.7},
                {fill=blue!60,draw=black,fill opacity=0.7},
                {fill=green!60,draw=black,fill opacity=0.7},
            },
            height=7cm,  % Adjust the height to reduce the overall height
        ]
        \addplot+[error bars/.cd, y dir=both, y explicit] coordinates {
           (1.25, 15193.21) +- (820.11, 820.14)
           (3.75, -435.61) +- (2842.07, 2215.44)
        };

        \addplot+[error bars/.cd, y dir=both, y explicit] coordinates {
            (6.25, 13816.30) +- (1221.50, 1000.60)
            (8.75, -9752.65) +- (3414.23, 2395.22)
        };

        \addplot+[error bars/.cd, y dir=both, y explicit] coordinates {
            (11.25, 19991.47) +- (8183.39, 8959.12)
            (13.75, 11539.83) +- (11262.54, 6914.32)
            (16.25, 35358.31) +- (8806.37, 3194.86)
        };       
        \legend{TS1, TS2, TS3}
        \end{axis}
    \end{tikzpicture}
    \caption{Average profit for TS1, TS2, \& TS3 in each market. Error bars highlight the results for each quantile pair.}
    \label{fig:marketcomparison}
\end{figure}

We assess three distinct trading strategies: TS1, TS2, and TS3, which employ quantile forecasts and trading constraints involving a hypothetical battery with a capacity $B_{c}$ set as 1 MWh, 80\% discharge efficiency ($\mathcal{E}{d}$), and 98\% charge efficiency ($\mathcal{E}{c}$). Each strategy is evaluated over a period of 12 months, covering both the DAM and BM. 
TS3 emerges as the top-performing strategy, surpassing TS1 and TS2 in the DAM. However, both TS1 and TS2 encounter challenges in the BM, as illustrated in Figure \ref{fig:marketcomparison}, which depicts the profit performance of each strategy over the evaluation period. TS3's superiority is attributed to its flexible timestamp orders and increased trading frequency, which are crucial in quantile trading and allow it to capitalize on market fluctuations effectively.
The inclusion of TS3-Dual, simultaneously participating in both DAM and BM, emphasizes the advantages of market diversification in energy trading, mitigating risks associated with market volatility. The observed variance in Figure \ref{fig:marketcomparison} across trading strategies for each quantile pair stems from differing trade execution numbers and market conditions. TS3's closely aligned quantile pairs result in more identified trades and increased profit opportunities, contrasting with TS1, which, with minimal variance in quantile pairs (Figure \ref{fig:marketcomparison}), identifies fewer distinctive trading opportunities.
Table \ref{table:MVES} furnishes a comprehensive analysis of the financial profits derived from TS3, utilizing a high-frequency strategy, across different quantile pairs and markets over the evaluation period. This breakdown portrays the varied and nuanced outcomes of each trading strategy, including their performance relative to key benchmarks such as perfect foresight (PF), signifying the theoretical maximum profit achievable with complete knowledge of future price movements, and our baseline quantile pair 0.5-0.5, serve to underscore the enhanced performance of TS3. Notably, despite its overall success, TS3 faces challenges in achieving profitability in the BM, reflecting the inherent unpredictability and volatility of this market.

\subsubsection*{Trading in the day-ahead market} 
Our analysis of trading strategies in the DAM reveals incremental improvements when compared to the baseline 0.5-0.5 quantile pair. For strategy TS3, the baseline quantile pair 0.5-0.5 consistently outperforms other quantile pairs in all four models, securing approximately 83\% of the potential profit of \euro34,703. Notably, the next most successful quantile, 0.3-0.5, attains approximately 63\% of the potential profit, as reported in Table \ref{table:MVES}.
The 0.5-0.5 quantile pair consistently attains the highest or closely approximates the top quantile pair, resulting in the highest average profitability across all trading strategies in the DAM. This indicates a limited impact for quantile trading in the DAM due to the relatively low error rates. 
% Furthermore, in the context of the selected trading strategies, it is clear that TS3 consistently outperforms TS1 and TS2, as shown in Figure \ref{fig:marketcomparison}. The buy/sell constraints in TS1 \& TS2 restrict profit potential.

\begin{sidewaystable*}
\centering
\setlength{\tabcolsep}{3pt}
\begin{tabular}{r@{}rr|crrrrrrr|c@{}|c@{}}
\toprule
& \multicolumn{11}{c}{Financial Profits from TS3: Bottleneck-Controlled Strategy} \\
\cmidrule{1-13}
& \textbf{Quantiles $\alpha$} 
& & & 0.5-0.5 & 0.1-0.3 & 0.3-0.5 & 0.5-0.7 & 0.7-0.9 & 0.3-0.7 & 0.1-0.9 & PF & Model Average \\
\midrule
\textbf{Market} & \textbf{Model} & &  & & & & & & & & & \\

DAM & KNN & &  & \textbf{\euro24,921} & \euro9,834 & \euro11,246 & \euro8,433 & \euro6,855 & \euro5,614 & \euro2,007 & \euro34,703 & \euro9,844\\
 & RF & &  & \textbf{\euro32,897} & \euro19,125 & \euro26,440 & \euro25,507 & \euro24,230 & \euro18,355 & \euro14,222 & \euro34,703 & \euro22,968 \\
 & LGBM & &  & \textbf{\euro33,236} & \euro26,143 & \euro29,067 & \euro32,056 & \euro32,769 & \euro29,673 & \euro23,364 & \euro34,703 & \textbf{\euro29,472} \\
 & DNN & &  & \textbf{\euro24,745} & \euro16,920 & \euro21,249 & \euro20,209 & \euro15,964 & \euro17,029 & \euro7,638 & \euro34,703 & \euro17,679 \\
\midrule

BM & KNN & &  & \euro-4,185 & \euro3,294 & \textbf{\euro6,400} & \euro6,139 & \euro5,366 & \euro-12 & \euro0 & \euro217,326 & \euro2,428 \\
 & RF & &  & \euro25,163 & \euro17,566 & \euro26,770 & \textbf{\euro27,042} & \euro20,641 & \euro13,562 & \euro-215 & \euro217,326 & \textbf{\euro18,647} \\
 & LGBM & &  & \euro16,547 & \euro13,333 & \euro18,549 & \textbf{\euro20,314} & \euro17,708 & \euro8,355 & \euro0 & \euro217,326 & \euro13,544 \\
 & DNN & &  & \euro7,253 & \euro8,240 & \euro15,458 & \textbf{\euro20,320} & \euro17,252 & \euro10,921 & \euro1,325 & \euro217,326 & \euro11,538 \\
\midrule

DUAL & KNN & &  & \euro32,517 & \euro32,156 & \euro33,429 & \textbf{\euro34,501} & \euro31,668 & \euro29,851 & \euro12,413 & \euro70,389 & \euro29,505 \\
 & RF & &  & \textbf{\euro42,979} & \euro37,600 & \euro40,781 & \euro41,629 & \euro42,401 & \euro38,431 & \euro31,094 & \euro70,389 & \textbf{\euro39,274} \\
 & LGBM & &  & \euro39,001 & \euro36,996 & \euro39,244 & \euro41,311 & \textbf{\euro41,619} & \euro38,299 & \euro36,055 & \euro70,389 & \euro38,932 \\
 & DNN & &  & \euro34,511 & \euro33,503 & \euro35,273 & \textbf{\euro36,769} & \euro35,024 & \euro34,319 & \euro26,644 & \euro70,389 & \euro33,720 \\
\midrule

DAM & Average $\alpha$ & &  & \textbf{\euro28,950} & \euro18,006 & \euro22,001 & \euro21,551 & \euro19,954 & \euro17,667 & \euro11,808 & \euro34,703 & \euro19,991 \\
BM & Average $\alpha$ & &  & \euro11,194 & \euro10,608 & \euro16,794 & \textbf{\euro18,454} & \euro15,242 & \euro8,206 & \euro277       & \euro217,326 & \euro11,539 \\
DUAL & Average $\alpha$ & &  & \euro37,252 & \euro35,064 & \euro37,182 & \textbf{\euro38,553} & \euro37,678 & \euro35,225 & \euro26,551 & \euro70,389 & \textbf{\euro35,358} \\
\bottomrule
\end{tabular}
\caption{Financial Profits from TS3: Bottleneck-Controlled Strategy}
\label{table:MVES}
\end{sidewaystable*}

\subsubsection*{Trading in the balancing market} 
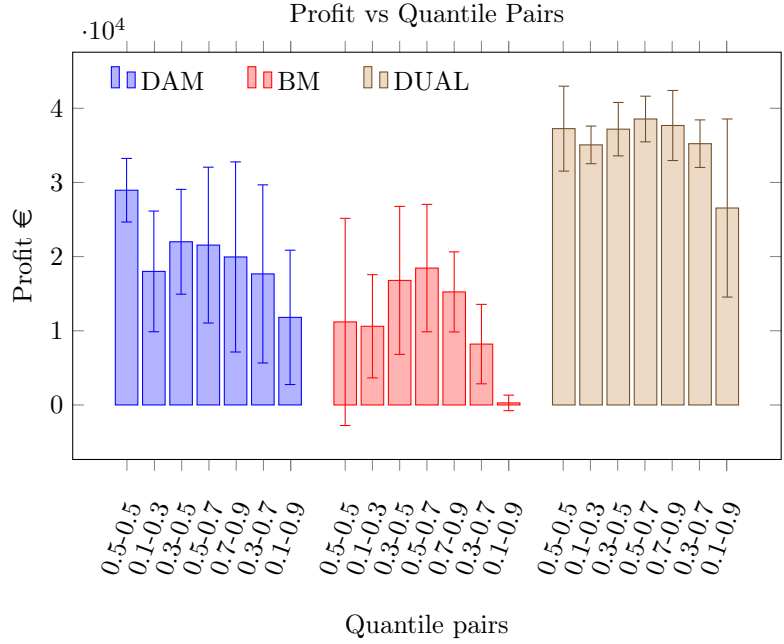
\begin{figure}[ht!]
\centering
    \begin{tikzpicture}
        \begin{axis}[
            ybar,
            bar width=0.3cm,
            ylabel={Profit \euro},
            xtick={-0.01, 1.0, 2.0, 3.0, 4.0, 5.0, 6.0, 6.0,
                   8.0, 9.0, 10.0, 11.0, 12.0, 13.0, 14.0, 14.0, 
                   16.0, 17.0, 18.0, 19.0, 20.0, 21.0, 22, 22.0}, 
            xticklabels={0.5-0.5, 0.1-0.3, 0.3-0.5, 0.5-0.7, 0.7-0.9, 0.3-0.7, 0.1-0.9, ,
                         0.5-0.5, 0.1-0.3, 0.3-0.5, 0.5-0.7, 0.7-0.9, 0.3-0.7, 0.1-0.9, ,
                         0.5-0.5, 0.1-0.3, 0.3-0.5, 0.5-0.7, 0.7-0.9, 0.3-0.7, 0.1-0.9, },
            xticklabel style={text width=1.5cm, align=center, rotate=70, anchor=north east},
            enlarge x limits=0.15,
            title=Profit vs Quantile Pairs,
            xlabel={Quantile pairs},
            legend style={at={(0.58,0.98)}, anchor=north east, legend columns=-1, draw=none, /tikz/every even column/.append style={column sep=0.5cm},
            },
            % cycle list={ % Define distinct colors for bars
            %     {fill=yellow!60,draw=black,fill opacity=0.7},
            %     {fill=red!60,draw=black,fill opacity=0.7},
            %     {fill=blue!60,draw=black,fill opacity=0.7},
            % },
            height=7cm,  % Adjust the height to reduce the overall height
        ]
        \addplot+[error bars/.cd, y dir=both, y explicit] coordinates {
            (1, 28950.5925) +- (4204.6725, 4286.2475)
            (2, 18006.0775) +- (8171.5975, 8137.1825)
            (3, 22001.02) +- (7754.14, 7066.34)
            (4, 21551.8125) +- (10118.7225, 10504.8175)
            (5, 19954.7775) +- (13099.6075, 12814.5825)
            (6, 17667.9275) +- (12053.8975, 12005.4025)
            (7, 11808.0825) +- (9000.5325, 9055.9675)
        };

        \addplot+[error bars/.cd, y dir=both, y explicit] coordinates {
            (8, 11194.68) +- (15380.46, 13968.96)
            (9, 10608.7525) +- (7314.4725, 6958.1975)
            (10, 16794.845) +- (10394.005, 9975.765)
            (11, 18454.15) +- (8315.08, 8588.13)
            (12, 15242.3275) +- (7875.6375, 5399.2425)
            (13, 8206.755) +- (6219.375, 5355.905)
            (14, 277.285) +- (1093.145, 1047.715)
        };

        \addplot+[error bars/.cd, y dir=both, y explicit] coordinates {
            (15, 37252.68) +- (4735.22, 5727.3)
            (16, 35064.2375) +- (2507.6875, 2536.1125)
            (17, 37182.1575) +- (3752.9675, 3599.4325)
            (18, 38553.1675) +- (3051.7375, 3076.8025)
            (19, 37678.3825) +- (6010.2825, 4723.0875)
            (20, 35225.585) +- (4373.905, 3205.875)
            (21, 26551.9325) +- (12138.3725, 12003.5675)
        };

        \legend{DAM, BM, DUAL}
        \end{axis}
    \end{tikzpicture}
    \caption{TS3 Profit for DAM, BM \& DUAL quantile pairs. The error bars are relative to the different forecasting models.}
    \label{fig:QP}
\end{figure}

Quantile-based strategies gain greater importance in the inherently unpredictable BM. Within TS3, as detailed in Table \ref{table:MVES}, the 0.5-0.5 quantile averages \euro11,194, approximately 5\%, out of \euro217,326, while the 0.5-0.7 quantile achieves \euro18,454, approximately 8.5\%. In the context of the BM, the 0.5-0.7 quantile pair emerges as the top-performing choice for three out of four models, closely followed by the 0.7-0.9 quantile pair. In addition, significant variance exists among the quantile pairs across the three trading strategies, as illustrated in Figure \ref{fig:QP}. This variability is more closely linked to the performance of the chosen trading strategy rather than the specific quantile selection.
These findings highlight the critical role of both quantile pair selection and high-frequency trading in the BM. TS3, characterized by its adaptability and increased trading frequency, excels.
In contrast to the DAM, where profits closely align with the PF benchmark due to its low error rate, the BM falls short of reaching the profitability achieved with a PF of \euro217,326. This discrepancy underscores the challenges associated with accurately forecasting the BM, given its inherent unpredictability in a highly volatile market with numerous predictors. Nevertheless, it underscores the untapped profit potential for market participants with successful, precise forecasting methodologies.

\subsubsection*{Trading Dual Market positions} 
The trading strategy TS3-Dual focuses on both markets simultaneously, aligning DAM positions with BM positions. As detailed in Table \ref{table:MVES}, the deployment of the 0.5-0.5 quantile within this strategy results in a profit of \euro37,252, 53\% of the potential profit \euro70,389, while the 0.5-0.7 quantile pair achieves an average of 55\% with \euro38,553 in profit, it becomes clear as the top-performing choice, as reflected in Table \ref{table:MVES}.
Although the Dual Market's potential profit of \euro70,389 (32\%) falls short of the BM's potential profit of \euro217,326, the average return from TS3-Dual amounts to \euro35,358, outperforming the DAM return by 76\% and more than tripling the average BM return. 
Ultimately, the TS3-Dual approach consistently outpaces that of individual market trading across all strategies, as evidenced in Figure \ref{fig:marketcomparison}.

\subsubsection*{Performance of Non-Conventional Quantile Pairs} 
In this section, we evaluate the performance of non-conventional quantile pairs (0.3-0.5, 0.5-0.7, 0.7-0.9) in contrast to conventional quantile pairs (e.g., 0.5-0.5, 0.3-0.7, 0.1-0.9) within our trading strategy, with a specific focus on strategy TS3, which capitalizes on increased trading frequency.
Our analysis underscores the exceptional performance of non-conventional quantile pairs, particularly the 0.5-0.7 pair, in the context of strategy TS3. In the DAM, the 0.3-0.5 and 0.5-0.7 quantile pairs closely trail the baseline 0.5-0.5 pair. However, in the BM, where trading frequency plays a pivotal role, 0.3-0.5 and 0.5-0.7 emerge as the top-performing quantile pairs.
These findings underscore the critical importance of carefully aligning non-conventional quantile pairs with specific trading strategies and markets. Broader quantile pair ranges, such as 0.1-0.9, consistently lead to fewer trades and limited profitability, highlighting the importance of high-frequency trading in trading success.

\subsubsection*{Model Selection and Forecast Accuracy}
In this section, we evaluate the performance of four distinct models—KNN, RF, LGBM, and DNN—in shaping our trading decisions and profitability within the dynamic energy markets. 
The performance of these models is crucial for understanding their effectiveness in different market scenarios. Figure \ref{fig:MP} presents a visual comparison of the profits generated by each model for the DAM, BM, and DUAL scenarios. 
The LGBM and RF models excel in both the DAM and the BM, as depicted in Figure \ref{fig:MP}. In contrast, the DNN and KNN models struggle, leading to reduced profits in all market scenarios, particularly in strategy TS3.
Furthermore, to assess the accuracy of our forecasts, we examine the pinball scores associated with each model. Figure \ref{fig:pinball} illustrates the relationship between pinball scores and profitability for our trading strategy. Lower pinball scores indicate more accurate forecasts, which tend to result in higher profits. 
The profit-to-error ratios highlight the importance of accurate quantile-based forecasts in achieving success in electricity trading. Specifically, the RF model emerges as the top-performing choice, with the lowest error and the highest average profits across all market scenarios. This underscores the pivotal role of accurate quantile-based forecasts in addressing market-specific challenges and seizing opportunities.
These findings have practical implications for energy traders, emphasising the importance of precise quantile-based forecasts for profitable outcomes.

\begin{figure}[ht!]
\begin{minipage}{0.48\textwidth}
    \centering
    \begin{tikzpicture}[scale=0.8]
        \begin{axis}[
            xlabel={Error},
            ylabel={Profit \euro},
            grid=both,
            grid style={line width=0.1pt, draw=gray!20},
            major grid style={line width=0.2pt, draw=gray!50},
            legend style={at={(.99999,.995)}, anchor=north east,  draw=none, /tikz/every even column/.append style={column sep=0.5cm},
            },
            width=8cm,  % Adjust the width of the plot
            height=7cm,  % Adjust the height of the plot
        ]

    \addplot coordinates {
        (5.69, 9844.72428571429)
        (4.73, 17679.6885714286)
        (2.67, 22968.4914285714)
        (2.434, 29472.9757142857)
    };
    \addlegendentry{DAM}

    \addplot coordinates {
        (14.09, 2428.92571428571)
        (13.16, 11538.7528571429)
        (12.49, 13544.2257142857)
        (12.16, 18647.4071428571)
    };
    \addlegendentry{BM}

    \addplot coordinates {
        (9.89, 29505.4242857143)
        (8.945, 33720.9414285714)
        (7.462, 38932.71)
        (7.415, 39274.1485714286)
    };
    \addlegendentry{DUAL} 
        \end{axis}
    \end{tikzpicture}
    \caption{TS3 Aggregate Pinball Score of Model vs Profit}
    \label{fig:pinball}
\end{minipage}
\hspace{0.5cm}
\begin{minipage}{0.48\textwidth}
    \centering
    \begin{tikzpicture}[scale=0.8]
        \begin{axis}[
            ybar,
            bar width=0.3cm,
            ylabel={Profit \euro},
            xtick={0.2, 1.2, 2.2, 3.2,       5, 6, 7, 8,      9.8, 10.8, 11.8, 12.8},
            xticklabels={KNN, DNN, LGBM, RF, KNN, DNN, LGBM, RF,  KNN, DNN, LGBM, RF},
            xticklabel style={text width=1.5cm, align=center, rotate=70, anchor=north east},
            enlarge x limits=0.15,
            xlabel={Model},
            legend style={at={(0.79,0.99)}, anchor=north east, legend columns=-1, draw=none, /tikz/every even column/.append style={column sep=0.5cm},
            },
            width=8cm,
            height=6cm
            ]

        \addplot+[error bars/.cd, y dir=both, y explicit] coordinates {
            (1, 9844.72428571429) +- (15077.1457142857, 7837.17428571429)
            (2, 17679.6885714286) +- (7066.2314285714, 10041.1885714286)
            (3, 29472.9757142857) +- (3763.86428571429, 6108.9257142857)
            (4, 22968.4914285714) +- (9929.2485714286, 8746.2614285714)
        };

        \addplot+[error bars/.cd, y dir=both, y explicit] coordinates {
            (5, 2428.92571428571) +- (3971.91428571429, 6614.70571428571)
            (6, 11538.7528571429) +- (8781.5571428571, 9213.7528571429)
            (7, 13544.2257142857) +- (6770.7142857143, 7544.2257142857)
            (8, 18647.4071428571) +- (8394.8728571429, 8863.2671428571)
        };

        \addplot+[error bars/.cd, y dir=both, y explicit] coordinates {
            (9, 29505.4242857143) +- (17091.8642857143, 4996.0057142857)
            (10, 33720.9414285714) +- (3048.4785714286, 3076.4914285714)
            (11, 38932.71) +- (1705.8314285714, 2079.9285714286)
            (12, 39274.1485714286) +- (2905.8314285714, 2979.9285714286)
        };
     
        \legend{DAM, BM, DUAL}
        \end{axis}
    \end{tikzpicture}
    \caption{TS3 Profit for Models in the DAM, BM \& Dual with Range in Quantile Pair}
    \label{fig:MP}
\end{minipage}
\end{figure}

\subsection{Economic Viability - Scenario Analysis}\label{EconVia}
In this section, we present a comprehensive analysis of the economic viability of BESS configurations, exploring various factors that influence their returns and performance. Our study encompasses a scenario analysis focusing on key elements such as construction costs, maintenance expenses, degradation-related costs, efficiency losses, and market dynamics. The goal is to evaluate the profitability and economic outlook of different BESS units over a potential lifespan of 10-15 years.
In the following discussion, as detailed in \cite{Baxter_Byrne, invinityWhatDoes}, we explore key factors influencing the returns derived from operating a BESS. These include:
\begin{itemize}
    \item CAPEX: Battery purchase, including ancillary systems, installation, commissioning, permits, and civil works.
    \item Fixed O\&M (\euro kW-year): Fixed maintenance costs and variable costs throughout the system's lifespan, based on operation hours or cycles. All batteries are modelled to have maintenance costs increasing at a rate of 2\% per year, as per the Sandia report \cite{sandiaUSEnergy}.
    \item Degradation-related costs: Expenses due to cycle-induced degradation, potentially requiring cell addition or replacement, leading to reduced battery capacity and efficiency over time.  Battery degradation is estimated at 1.55\% in line with lithium-ion battery data from \cite{graf2022drives}.
    \item Efficiency costs: Costs related to round-trip efficiency losses (\euro/kWh). All battery options are modelled with a 95\% charge \& discharge efficiency.
    \item Depth of Discharge: Energy discharged as a percentage of rated capacity.
    \item Ramp Rate: Ratios of rated energy to rated power, e.g. 2-hour Tesla megapack, 1-hour lithium-ion battery in Sandia report \cite{sandiaUSEnergy}.
    \item Calendar Life (years): Maximum lifespan from factors such as state of charge, e.g., Li-ion batteries end life below 60\% rated energy. All batteries have an assumed lifespan of $\sim$10-15 years.
    \item Market registration costs for trading in ISEM DAM and BM include an annual total of \euro18,294. These costs consist of a once-off entry fee, annual subscription fees, BM accession fee, BM participation fee, and variable trading fees, which apply per MWh traded. For a detailed fee breakdown, see \cite{semopxSEMOpxFees} for ISEM DAM \& IDM and \cite{semoJoiningBalancing} for BM participation. 
\end{itemize}

Items we will not factor in for issues on acquiring accurate data include:
\begin{itemize}
    \item Warranty: (\euro/kWh-year): Annual fees for quality assurance.
    \item Insurance (\euro/kWh): Premiums against unforeseen risks.
    \item End-of-life costs: Expenses for disassembly, transportation to recycling facilities, and safe disposal of lithium-ion cells.
\end{itemize}

\subsubsection*{Battery Options}
We analyze various battery options, utilising configurations provided in \cite{SEMC2021}, including smaller batteries with capacities of 3MWh (Scottish Power, Battery A) and 3.9 MWh (Tesla Megapack, Battery B) and larger batteries with capacities of 10MWh (Avolta storage, Battery C) and 39 MWh (10 Tesla Megapacks, Battery D). Revenue data for all batteries is derived from the leading dual-market model in TS3-Dual, utilizing configurations specific to the selected battery setup.

\subsubsection*{Smaller Batteries: A \& B}
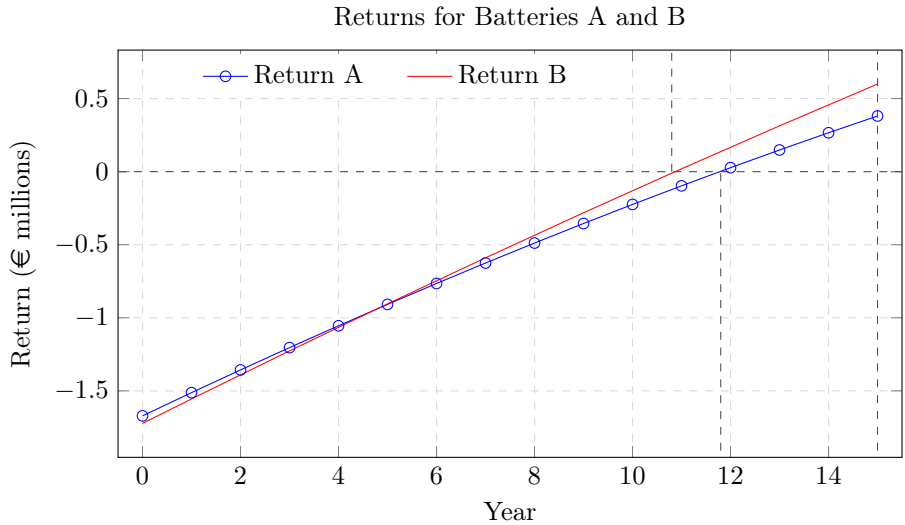
\begin{figure} % Start a figure environment
\begin{tikzpicture}
\begin{axis}[
    xlabel=Year,
    ylabel=Return (\euro $ $  millions),
    title=Returns for Batteries A and B,
    grid=both,
    legend style={
        at={(0.59,0.99)},
        anchor=north east,
        legend columns=-1,
        draw=none,
        /tikz/every even column/.append style={column sep=0.5cm},
    },
    width=.99\textwidth,  % Adjust the width
    height=7cm,  % Adjust the height
    extra y ticks={0},  % Add extra ticks on the y-axis
    extra y tick style={grid=none, tick label style={font=\tiny}},
    extra y tick labels={},  % Empty labels for the extra ticks
    grid style={dashed, gray!30},  % Dashed grid lines
    xmin=-0.5,  % Set the minimum value on the x-axis
    xmax=15.5,  % Set the maximum value on the x-axis
]

\addplot[mark=o, blue] table {
    Year    Return A
    0       -1.671000
    1       -1.512178
    2       -1.356493
    3       -1.203948
    4       -1.054546
    5       -.908292
    6       -.765187
    7       -.625234
    8       -.488433
    9       -.354785
    10      -.224290
    11      -.096947
    12      .027246
    13      .148291
    14      .266191
    15      .380951
};

\addplot[mark=s, red] table {
    Year    Return B
    0       -1.722628
    1       -1.555077
    2       -1.390680
    3       -1.226442
    4       -1.065412
    5       -.904547
    6       -.746944
    7       -.589513
    8       -.435395
    9       -.281456
    10      -.130882
    11      .019505
    12      .166477
    13      .313256
    14      .456570
    15      .599682
};

\legend{Return A, Return B}
% Add dark dashed lines at the 10 and 15-year marks
% \draw[dashed, black!70] (axis cs:10,-3000000) -- (axis cs:10,900000);
\draw[dashed, black!70] (axis cs:15,-30) -- (axis cs:15,9);
\draw[dashed, black!70] (axis cs:-0.5,0) -- (axis cs:15.5,0); 
\draw[dashed, black!70] (axis cs:10.8, 0) -- (axis cs:10.8, 90); % Vertical dashed line at x=10.8
\draw[dashed, black!70] (axis cs:11.8, -30) -- (axis cs:11.8, 0); % Vertical dashed line at x=11.8
\end{axis}
\end{tikzpicture}
\caption{15-year cost breakdown for the 3 MW Scottish Power and 3.9 MW Tesla capacity}
\label{smallbatt}
\end{figure} % End the figure environment

\begin{itemize}
    \item Battery A, Scottish Power (3 MWh) - This battery features a 1-hour cycle, a maximum discharge capacity of 3 MWh, and operates with a charge/discharge efficiency of 95\%. With a construction cost of \euro1,671,000.00 (\euro557 per kWh) and annual maintenance expenses of \euro11,000 (\euro4 per kWh), it represents a compact and efficient option suitable for applications with shorter duration requirements.

    \item Battery B, Tesla Megapack (3.9 MWh) - Designed for a 2-hour cycle, Battery B shares similar charge/discharge efficiencies and ramp rates with Battery A. Its purchase costs amount to \euro1,722,628.98 (\euro447 per kWh), and annual maintenance costs are \euro7,730.96 (\euro2 per kWh). The extended capacity makes it well-suited for applications demanding longer discharge periods.

    \item Battery C, Avolta Storage (10 MWh) - This battery is configured with a 1-hour cycle, a maximum discharge capacity of 9 MWh, and a charge/discharge efficiency of 95\%. With a construction cost of \euro4,850,000 (\euro485 per kWh) and annual maintenance costs of \euro100,000 (\euro10 per kWh), designed to balance capacity and efficiency, catering to applications with larger scale and short duration requirements.

    \item Battery D, Tesla Megapack*10 (38.5 MWh) - A larger-scale option, features a 2-hour cycle and shares similar charge/discharge efficiencies and ramp rates with Battery B. The purchase cost, accounting for a bulk discount for ordering 10 units, is \euro13,726,186.17 (\euro356 per kWh), with annual maintenance costs of \euro61,593.89 (approximately \euro2 per kWh). This configuration, suitable for substantial energy storage needs, highlights the potential economic benefits of scaling up BESS installations.
\end{itemize}

\begin{figure} % Start a figure environment
\begin{tikzpicture}
\begin{axis}[
    xlabel=Year,
    ylabel=Return (\euro $ $  millions),
    title=Returns for Batteries C and D,
    grid=both,
    legend style={
        at={(0.59,0.99)},
        anchor=north east,
        legend columns=-1,
        draw=none,
        /tikz/every even column/.append style={column sep=0.5cm},
    },
    width=.99\textwidth,  % Adjust the width
    height=7cm,  % Adjust the height
    extra y ticks={0},  % Add extra ticks on the y-axis
    extra y tick style={grid=none, tick label style={font=\tiny}},
    extra y tick labels={},  % Empty labels for the extra ticks
    grid style={dashed, gray!30},  % Dashed grid lines
    xmin=-0.5,  % Set the minimum value on the x-axis
    xmax=15.5,  % Set the maximum value on the x-axis
]

\addplot[mark=^, green] table {
    Year    Return C
    0       -4.850000
    1       -4.436707
    2       -3.829655
    3       -3.436882
    4       -2.842186
    5       -2.470087
    6       -1.887564
    7       -1.536285
    8       -.965729
    9       -.635403
    10      -.076582
    11      0.232665
    12      0.780006
    13      1.068062
    14      1.604205
    15      1.870967
};

\addplot[mark=d, purple] table {
    Year    Return D
    0       -13.726186
    1       -11.412092
    2       -9.136336
    3       -6.898944
    4       -4.699931
    5       -2.539304
    6       -0.417062
    7       1.666806
    8       3.712322
    9       5.719514
    10      7.688420
    11      9.619090
    12      11.511582
    13      13.365962
    14      15.182308
    15      16.960707
};

\legend{Return C, Return D}
\draw[dashed, black!70] (axis cs:10.1,-20) -- (axis cs:10.1,0000000);
\draw[dashed, black!70] (axis cs:15,-20) -- (axis cs:15,20);
\draw[dashed, black!70] (axis cs:-0.5,0) -- (axis cs:15.5,0); 
\draw[dashed, black!70] (axis cs:6.2, 0) -- (axis cs:6.2, 20); 
\end{axis}
\end{tikzpicture}
\caption{15-year cost breakdown for Avolta Storage (C) and Tesla MegaPacks (D) totaling 38.5 MWh}
\label{biggbatt}
\end{figure}
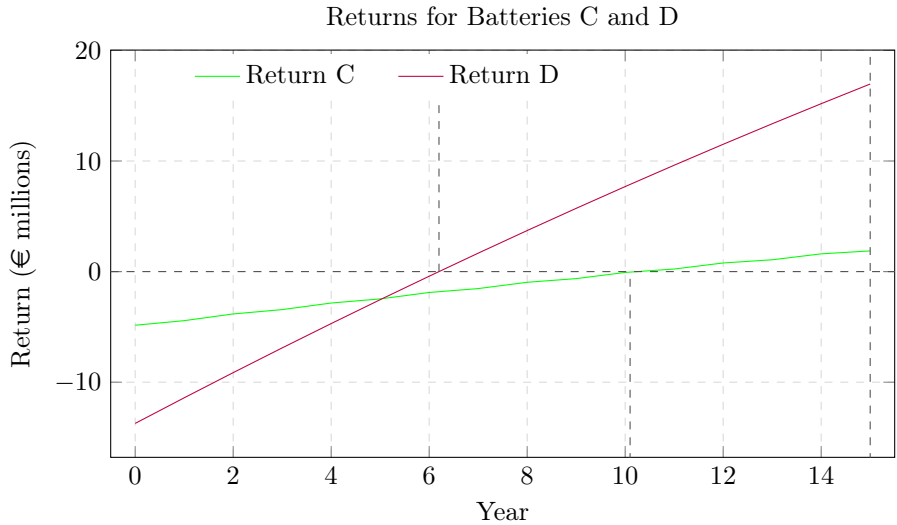 % End the figure environment

\subsubsection*{Economic Viability of BESS Units} 
This section assesses the economic viability of various BESS configurations, looking at what year they are profitable and their performance over a potential battery life of 10-15 years. Figure \ref{smallbatt} illustrates the 15-year return on investment for batteries A and B. Figure \ref{biggbatt} compares the 15-year returns for batteries C and D.
\begin{enumerate}
    \item A 3 MW Scottish Power unit with construction costs of \euro557/kW and maintenance costs of \euro11/kW, doesn't achieve profitability until year 12, with a a 16\% return by year 15.

    \item A 3.9 MW Tesla Megapack with construction costs of \euro447/kW and maintenance costs of \euro2/kW, doesn't achieve profitability until year 11, with a a 34\% return by year 15.

    \item A 10 MW Avolta Storage unit with construction costs of \euro485/kW and maintenance costs of \euro10/kW, achieves profitability by year 11, with a 38\% return by year 15.

    \item A 38.5 MW Tesla Megapacks unit with construction costs of \euro356/kW and maintenance costs of \euro2/kW, achieves profitability by year 7, a return of 56\% by year 10 and 123\% by year 15. The results emphasize the impact of size and cost dynamics on BESS economic outlook.

\end{enumerate}
The findings highlight the evolving landscape of BESS and the crucial role of cost reduction. When considering potential interest in construction and maintenance costs, Batteries A and C point towards alternative investment opportunities. In contrast, Batteries B and, notably, Battery D exemplify economic viability. These results underscore the recent advancements in BESS cost-effectiveness and emphasize the significance of appropriate battery sizing.
They reveal the intricate link between construction and maintenance expenses, battery size, and long-term profitability, shaping the future of energy storage technologies and their grid integration.

\section{Conclusion}\label{conclusionsec}
In this study, we conducted an analysis of BESS trading, employing quantile forecasting in both the DAM and BM. Our investigation encompassed trading strategies, constraints, and the impact of quantile forecasts on profitability in BESS trading. 
Our findings highlight the effectiveness of dual-market strategies, with TS3-Dual emerging as the most successful, with an emphasis on adaptability and high-frequency trading. In the DAM, the baseline 0.5-0.5 quantile pair consistently outperformed other quantile pairs, while in the BM, the 0.5-0.7 quantile pair exhibited the best performance, emphasizing the critical role of precise quantile pair selection and trading frequency.
The choice of models also significantly impacted profitability, with RF models excelling in both the DAM and BM. Low pinball scores were associated with increased profitability, emphasizing the importance of accurate quantile-based forecasts. 
In terms of the economic viability of BESS trading, analyzing cost implications, we demonstrate the potential for profitable engagement in energy markets with appropriate battery sizing and trading strategies. Future research could explore the incorporation of intra-day markets into trading strategies, delve into the price impact on BM trading, and investigate the utilization of reinforcement learning techniques for trading (e.g., \cite{shavandi2022multi}). 
The promising results underscore the role of quantile forecasting and dual-market strategies in enhancing BESS profitability. Continued research and innovation in this domain could further amplify the success of BESS in the evolving energy landscape.

\section*{Acknowledgments}
This work was conducted with the financial support of Science Foundation Ireland under Grant Nos. 18/CRT/6223 and 12/RC/2289-P2 which are co-funded under the European Regional Development Fund. For the purpose of Open Access, the author has applied a CC BY public copyright licence to any Author Accepted Manuscript version arising from this submission.
\bibliography{mybibfile}

\end{document}